\documentclass{article}

\usepackage{arxiv}

\usepackage[utf8]{inputenc} 
\usepackage[T1]{fontenc}    
\usepackage{hyperref}       
\usepackage{url}            
\usepackage{booktabs}       
\usepackage{amsfonts}       
\usepackage{nicefrac}       
\usepackage{microtype}      
\usepackage{lipsum}
\usepackage{graphicx}
\graphicspath{ {./images/} }
\usepackage{amsmath} 

\title{A Hybrid Approach for Depression Classification: Random Forest-ANN Ensemble on Motor Activity Signals}

\author{\href{https://orcid.org/0009-0000-0619-4174}{Anket Patil}\thanks{} \\
	Department of Information Technology\\
	K. J. Somaiya Institute of Technology \\
	Maharashtra, India \\
	\texttt{patilanket11@gmail.com} \\
	\And
	\href{https://orcid.org/0009-0003-0525-5627}{Dhairya Shah} \\
	Department of Information Technology\\
	K. J. Somaiya Institute of Technology \\
	Maharashtra, India \\
	\texttt{dhairya.as@somaiya.edu} \\
        \And
	\href{https://orcid.org/0009-0006-0450-243X}{Abhishek Shah} \\
	Department of Information Technology\\
	K. J. Somaiya Institute of Technology \\
	Maharashtra, India \\
	\texttt{ahs1@somaiya.edu} \\
        \And
	\href{https://orcid.org/0000-0000-0000-0000}{Mokshit Gala} \\
	Department of Information Technology\\
	K. J. Somaiya Institute of Technology \\
	Maharashtra, India \\
	\texttt{mokshit.gala@somaiya.edu} \\
}

\begin{document}
\maketitle

\begin{abstract}
	Regarding the rising number of people suffering from mental health illnesses in today's society, the importance of mental health cannot be overstated. Wearable sensors, which are increasingly widely available, provide a potential way to track and comprehend mental health issues. These gadgets not only monitor everyday activities but also continuously record vital signs like heart rate, perhaps providing information on a person's mental state. Recent research has used these sensors in conjunction with machine learning methods to identify patterns relating to different mental health conditions, highlighting the immense potential of this data beyond simple activity monitoring. In this research, we present a novel algorithm called the Hybrid Random forest - Neural network that has been tailored to evaluate sensor data from depressed patients. Our method has a noteworthy accuracy of 80\% when evaluated on a special dataset that included both unipolar and bipolar depressive patients as well as healthy controls. The findings highlight the algorithm's potential for reliably determining a person's depression condition using sensor data, making a substantial contribution to the area of mental health diagnostics.
\end{abstract}


\section{Introduction}
Global health is greatly impacted by the widespread mental health problems of depression and bipolar disorder. The World Health Organization (WHO) estimates that over 264 million people worldwide suffer from depression, making it the main cause of disability [1]. Although bipolar disorder only affects 1\% to 2\% of the world's population, it has a significant negative impact on the affected population's quality of life and functional impairment [2]. These diseases have significant financial repercussions; depression and anxiety are estimated to cost US\$ 1 trillion annually in lost productivity [3]. With the introduction of on-body sensors, personal health monitoring has undergone a revolutionary change. Today's people use enormous amounts of data every day for a variety of goals, such as improving life quality, tracking their fitness levels, and changing unhealthy habits. This information includes continuous records of heart rate and activity levels, which have considerable promise in the field of psychiatry and go beyond the simple metrics of daily steps taken or calories burned. Growing emphasis has been paid to the complex association between activity data and a variety of mental health problems such mood swings, stress management, and social disengagement [4–5]. Since 2010, mental health issues—with depression leading the list of most common illnesses [6]–[8] have been the primary reason for years lived with disability worldwide. Depression presents a variety of difficulties in the physical, financial, and emotional spheres, which frequently result in problems at work and sick days [9]. The underlying etiology of these illnesses involves a complex combination of genetic, environmental, and social variables, with biological rhythm disruptions—often sparked by environmental disturbances—showing up in afflicted people as altered motor activity patterns [10]. By examining actigraph data to find patterns of motor activity suggestive of depressive and bipolar illnesses, this study aims to advance our understanding of these disorders. The study intends to clarify the distinctive motor activity patterns connected to various mood disorders using cutting-edge statistical and machine learning technologies, potentially permitting better diagnostic and curative approaches.

The main contributions of this paper are: 
\begin{enumerate}
  \item Our study introduces a groundbreaking Hybrid Random Forest – Neural Network model for depression classification, promising enhanced accuracy in mental health diagnosis.
  \item Our findings directly benefit clinical practice by enabling early depression detection, potentially improving patient outcomes in mental healthcare.
\end{enumerate}

The paper is organized as follows: Section 2 provides a comprehensive literature review, summarizing prior work and baseline algorithms. Section 3 presents the proposed approach, which focuses on the novel Algorithm. Section 4 discusses the implementation results, demonstrating the effectiveness of the proposed approach. Section 5 concludes the work and suggests directions for future research.






\section{Related Work}

The existing work on applications of mental healthcare field are presented in this section, which has been the base for this study.

\subsection{Baseline Algorithms}

The use of classification algorithms is vital in the effort to identify trends and draw conclusions from data, especially in the area of mental health. Understanding the underlying structure of the data and creating accurate predictions depend greatly on the classification method's objective, which is to group data into predetermined groups. There are countless categorization algorithms in the literature, each with their own advantages, presumptions, and application to various kinds of data.

\subsubsection{Artificial Neural Network [11]}

To convert neural networks into a classification or regression model in neural networks, utilize the activation function of the node in the output layer.

\begin{figure}[htbp]
    \centering
    \includegraphics[width=0.6\textwidth]{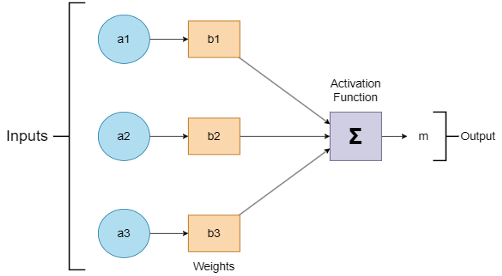}
    \caption{Artificial Neural Network Model}
    \label{fig:ann_model}
\end{figure}

A regression problem's output node has a linear activation function (or none at all). Continuous output from $-\infty$ to $+\infty$ is the range for a linear function. As a result, the output layer will be based on regression and be a function of the nodes in the layer before it, just as it was for testing the proposed system. Figure~\ref{fig:ann_model} depicts the ANN Model's system flow. In the context of the proposed study, the artificial neural network (ANN) model takes various activity statistics ($a_1$, $a_2$, $a_3$) as input and classifies the binary variable "state" which denotes sad vs. non-depressed states, by using linked layers of neurons.

\subsubsection{Random Forest [12]}

A random forest functions as a group of binary regression trees. An independent subset of variables is used to create these enormous numbers of binary regression trees. The variables for division are selected at random using Random Forest, and the decision trees are constructed using bootstrapped samples from the dataset.

\begin{figure}[htbp]
    \centering
    \includegraphics[width=0.6\textwidth]{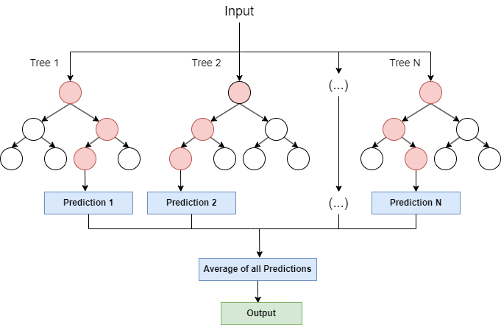}
    \caption{Random Forest Model}
    \label{fig:random_forest_model}
\end{figure}

Figure~\ref{fig:random_forest_model} depicts the Random Forest Regressor Model's system flow. For the system covered by the study, the random forest model applies activity statistics, such as mean logarithmic activity, standard deviation, minimum and maximum logarithmic activity, and proportion of zero activity, as input features to categorize the binary variable 'state' which denotes depressed or non-depressed states, through an ensemble of decision trees.

\subsection{Mental Health Monitoring Systems}

O’brien et al. [13] using a wrist-worn activity monitoring device to track persons with late-life depression, researchers discovered that their physical activity was lower than that of healthy controls.

In another study of Faurholt Jepsen et al. [14], they used smartphone sensors to track patients with bipolar illness and discovered that the more severe the depression symptoms, the fewer incoming calls were answered and the fewer outbound calls were recorded. Based on cell tower IDs, they also discovered that sad patients traveled less. Our work varies from the previous two in that we automatically categorize depressed vs. non-depressed individuals using machine learning on statistical variables generated from the sensor data.

ActiGraphs were used in a research to examine the viability of community-based remote biometric data gathering in people with behavioral signs of Alzheimer's disease [15]. The value of actigraphy baseline data (mean motor activity; MMA) was evaluated in the study. This study, although being specifically focused on Alzheimer's disease, highlights the potential and broader usefulness of actigraphy in monitoring motor activity across a range of mental and neurological diseases. The research reported in this paper tries to close this gap by developing a unique, diagnosis-focused ML model to determine whether or not a person is depressed.

\section{Proposed Approach}

\begin{figure}[htbp]
    \centering
    \includegraphics[width=0.6\textwidth]{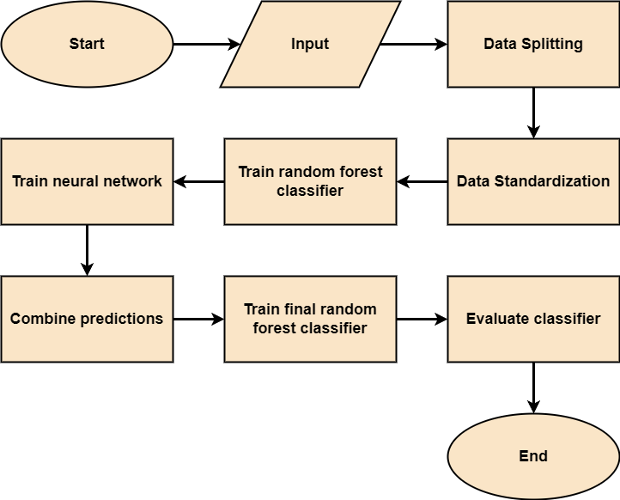}
    \caption{Architectural Design of the Proposed Approach}
    \label{fig:approach}
\end{figure}

The proposed approach for classifying depression is presented in this section with a comprehensive description of its architectural design illustrated in Figure~\ref{fig:approach}. Following this is a step-by-step breakdown of the various phases involved in this approach.

\subsection{Data Acquisition}

The acquired dataset is divided into two separate files that represent the control group and the condition group, respectively. Each participant's data is encapsulated, and the actigraph data amassed over a predetermined time period is chronicled. Each entry includes a timestamp, the measurement date, and the activity measurement obtained from the actigraph watch. The data is collected at one-minute intervals.

\subsection{Data Pre-Processing}

The activity column underwent a log transformation to lessen the impact of extreme values. The data was then divided into daily segments using a date-based filtering strategy, allowing for a controlled and methodical study of the motor activity patterns across time.

\subsection{Feature Extraction}

Several statistics, including the mean, standard deviation, minimum, maximum, and zero count of the log-transformed activity values, are retrieved in the code for each distinct date in the dataset. For each date, these retrieved characteristics are kept in a dictionary.

\subsection{Proposed Machine Learning Novel Algorithm: Hybrid Random Forest – Neural Network}

The Hybrid Random Forest – Neural Network combines the power of two distinct machine learning techniques, Random Forests (RF) and Artificial Neural Networks (ANN), to create a formidable ensemble model for predictive analytics. The algorithm is designed as follows:

\textbf{Step 1: Data Splitting}

Split the dataset into training and testing sets using an 80-20 split ratio.

\textbf{Step 2: Data Standardization}

Standardize the feature data using a StandardScaler to ensure mean=0 and standard deviation=1.

\textbf{Step 3: Random Forest Classifier}

\begin{enumerate}
    \item Initialize a Random Forest classifier with the following parameters:
    \begin{itemize}
        \item Number of decision trees (n estimators): 100
        \item Random seed (random state): 42
    \end{itemize}
    \item Train the Random Forest classifier on the standardized training data.
    \item Predict labels for the testing data using the trained Random Forest classifier.
\end{enumerate}

\textbf{Step 4: Neural Network Model}

\begin{enumerate}
    \item Define a feedforward neural network model with the following layers:
    \begin{itemize}
        \item Input layer: Shape matching the number of features.
        \item Hidden layer 1: 64 neurons with ReLU activation.
        \item Hidden layer 2: 32 neurons with ReLU activation.
        \item Output layer: 1 neuron with sigmoid activation for binary classification.
    \end{itemize}
    \item Compile the neural network with the following settings:
    \begin{itemize}
        \item Optimizer: Adam
        \item Loss function: Binary cross-entropy
        \item Metrics: Accuracy
    \end{itemize}
    \item Train the neural network on the standardized training data for 100 epochs with a batch size of 16.
\end{enumerate}

\textbf{Step 5: Combining Predictions}

Combine the binary predictions from the Random Forest classifier and the Neural Network.

\textbf{Step 6: Final Random Forest Classifier}

\begin{enumerate}
    \item Initialize a final Random Forest classifier with the following parameters:
    \begin{itemize}
        \item Number of decision trees (n estimators): 100
        \item Random seed (random state): 42
    \end{itemize}
    \item Train the final Random Forest classifier on the combined predictions obtained in Step 5 using the testing labels.
\end{enumerate}

\textbf{Step 7: Evaluation}

Evaluate the final classifier's performance using a classification report, which includes precision, recall, F1-score, and accuracy.

\section{Implementation Details}

In the ensuing section, we delineate the implementation intricacies and the resultant outcomes of the analytical procedures undertaken in this study.

\subsection{Dataset}

The dataset [16] has been divided into two separate sections, one of which represents the control group and the other of which represents the condition group. The actigraph data collected over a certain period is housed in a structured manner that contains each patient's data. The actigraph data includes activity measures obtained from an actigraph watch, timestamps at one-minute intervals, and the measurement date. Additionally, the dataset provides MADRS (Montgomery-Sberg Depression Rating Scale) scores, which include a unique identifier, the number of days of measurements, gender, age groups, the type of affective disorder (bipolar II, unipolar depressive, or bipolar I), the presence of melancholia, patient status (inpatient or outpatient), education level categorized in years, marital status, employment status, and MADRS scores for each patient.

\subsection{Algorithmic Parameters}

The implementation of Machine algorithms in this work involved careful selection and tuning of algorithmic hyperparameters. The algorithmic parameters used are depicted in Table~\ref{tab:algorithm_params}.

\begin{table}[htbp]
    \centering
    \caption{Algorithmic Parameters for Random Forest and Neural Network}
    \label{tab:algorithm_params}
    \renewcommand{\arraystretch}{1.5} 
    \begin{tabular}{|l|l|l|}
        \hline
        \textbf{Model} & \textbf{Description} & \textbf{Value} \\
        \hline
        Random Forest & Number of decision trees & 100 \\
                      & Maximum depth & None \\
                      & Minimum samples to split & 2 \\
                      & Minimum samples at leaf node & 1 \\
        \hline
        Neural Network & Learning Rate & 0.001 \\
                       & Number of Epochs & 100 \\
                       & Batch Size & 16 \\
                       & Standardization & Yes \\
                       & Hidden Layer 1 Units & 64 \\
                       & Hidden Layer 2 Units & 32 \\
        \hline
    \end{tabular}
\end{table}

\subsection{Evaluation Criteria}

To analyze the performance of the Machine Learning models, the following measures are used to evaluate the performance.

\subsubsection{Accuracy [17]}

In classification tasks, accuracy is a key parameter that measures the ratio of accurately predicted instances to all occurrences in the dataset. It offers a clear idea of the model's overall performance.

\begin{equation}
    \text{Accuracy} = \frac{\text{True Positives} + \text{True Negatives}}{\text{Total Instances}}
\end{equation}

\subsubsection{Precision [18]}

Precision is a measure of the accuracy of the positive predictions made by a classification model. It calculates the ratio of accurately predicted positive observations to all anticipated positive observations.

\begin{equation}
    \text{Precision} = \frac{\text{True Positives}}{\text{True Positives} + \text{False Positives}}
\end{equation}

\subsubsection{Recall [19]}

Recall, often referred to as Sensitivity or True Positive Rate, measures a model's capacity to find all pertinent instances within the data.

\begin{equation}
    \text{Recall} = \frac{\text{True Positives}}{\text{True Positives} + \text{False Negatives}}
\end{equation}

\subsubsection{F1 Score [20]}

The F1 Score provides a balance between precision and recall by calculating a harmonic mean of accuracy and recall.

\begin{equation}
    \text{F1 Score} = \frac{2 \times \text{Precision} \times \text{Recall}}{\text{Precision} + \text{Recall}}
\end{equation}
\begin{figure}[htbp]
    \centering
    \includegraphics[width=0.6\textwidth]{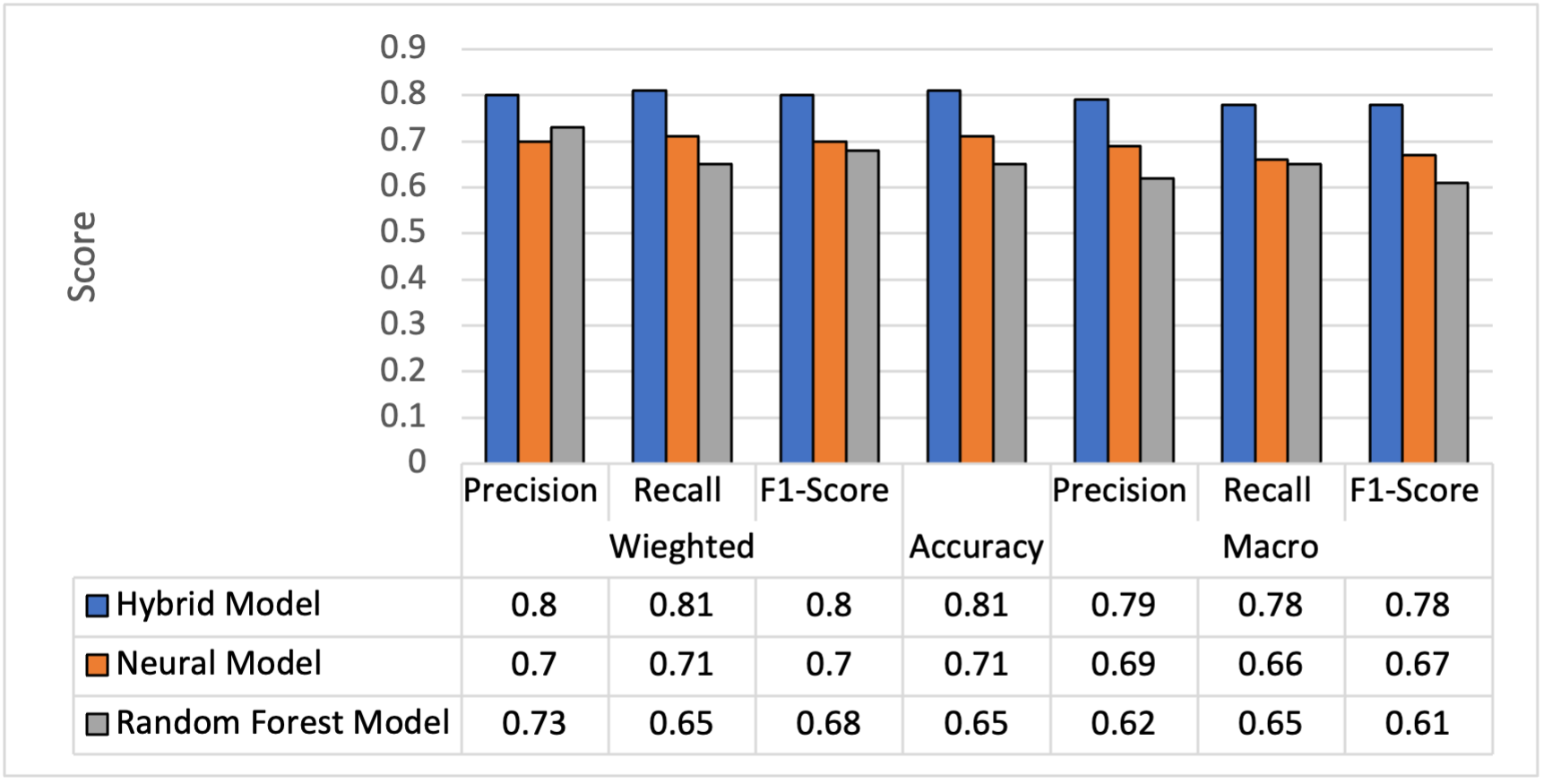}
    \caption{Performance Comparison of Machine Learning Algorithms for Classification}
    \label{chart:performance_comparison}
\end{figure}
\subsection{Results and Discussions}

The findings derived from the conducted research are presented in this section.

Results from binary classification models, including the Hybrid Model, Neural Model, and Random Forest Model, are displayed in Chart~\ref{chart:performance_comparison}. In comparison to its individual components, the hybrid model excels across all evaluation criteria. With an accuracy rating of 0.81, the Hybrid Model outperforms both the Neural Model (0.71) and the Random Forest Model (0.65) in terms of accuracy. This demonstrates the noteworthy accuracy of the Hybrid Model's predictions for about 81\% of the dataset. Additionally, the Hybrid Model performs admirably in terms of weighted precision, recall, and F1-score with scores of 0.8, 0.81, and 0.8, respectively, demonstrating its efficiency in categorizing cases across both classes while taking class imbalance into account. The Hybrid Model retains high performance in the macro-average (M) category, treating each class equally, with precision, recall, and F1-score all at 0.78, illustrating its constant prediction abilities without bias toward any particular class. The Neural Model and Random Forest Model, on the other hand, score significantly worse in both weighted and macro-average categories. The Hybrid Model is the best option for this binary classification problem since it can use the advantages of both models to produce a more reliable and accurate classification strategy.

\section{Conclusion}
The study presents a novel method for categorizing depression that is based on a Hybrid Random Forest-Neural Network model. Across all evaluation criteria, our hybrid model performs better than both the standalone Neural Network and Random Forest models. The Hybrid Model's astounding 81\% accuracy rate demonstrates its outstanding capacity for making accurate forecasts. Additionally, it showed excellent precision, recall, and F1-score, handling class imbalance and ensuring consistent predictive performance. This study highlights the ability of hybrid models to handle challenging tasks like depression categorization. This hybrid model provides a viable path for enhancing mental health diagnostics by integrating the benefits of both Random Forest and Neural Network techniques. It may help medical personnel identify patients earlier and intervene more effectively, thereby improving the quality of life for those who are depressed.
Future research should concentrate on determining how well the model generalizes across other datasets and populations. Key paths for additional research and practical application include investigating cutting-edge deep learning methods, merging real-time monitoring with wearable technologies, and performing clinical trials for real-world validation.

\bibliographystyle{unsrtnat}

\end{document}